\documentclass[10pt, a4paper]{article}

\usepackage[final]{template_files/lrec-coling2024} 
\usepackage{multirow}
\usepackage{booktabs}
\usepackage{tcolorbox}
\usepackage{xcolor}
\tcbuselibrary{most}
\usepackage{subfigure}
\usepackage{xspace}
\usepackage{float}
\usepackage{balance}

\title{ Multi-perspective Improvement of Knowledge Graph Completion with Large Language Models
}

\name{Derong Xu\textsuperscript{1,3}, Ziheng Zhang\textsuperscript{2}, Zhenxi Lin\textsuperscript{2}, Xian Wu\textsuperscript{2*}\thanks{* Corresponding authors. This work was done when Derong Xu was an intern at Tencent.}, Zhihong Zhu\textsuperscript{4},\\
\textbf{\large Tong Xu\textsuperscript{1*}, Xiangyu Zhao\textsuperscript{3*}, Yefeng Zheng\textsuperscript{2} and Enhong Chen\textsuperscript{1}}} 

\address{\textsuperscript{1}University of Science and Technology of China  \& State Key Laboratory of  Cognitive Intelligence, \\
 \textsuperscript{2}Jarvis Research Center, Tencent YouTu Lab, \textsuperscript{3}City University of Hong Kong, \textsuperscript{4}Peking University \\
         derongxu@mail.ustc.edu.cn, \{tongxu, cheneh\}@ustc.edu.cn,  zhihongzhu@stu.pku.edu.cn,\\
         \{zihengzhang, chalerislin, kevinxwu, yefengzheng\}@tencent.com, xianzhao@cityu.edu.hk}

\abstract{
Knowledge graph completion (KGC) is a widely used method to tackle incompleteness in knowledge graphs (KGs) by making predictions for missing links. Description-based KGC leverages pre-trained language models to learn entity and relation representations with their names or descriptions, which shows promising results. However, the performance of description-based KGC is still limited by the quality of text and the incomplete structure, as it lacks sufficient entity descriptions and relies solely on relation names, leading to sub-optimal results. To address this issue, we propose MPIKGC, a general framework to compensate for the deficiency of contextualized knowledge and improve KGC by querying large language models (LLMs) from various perspectives, which involves leveraging the reasoning, explanation, and summarization capabilities of LLMs to expand entity descriptions, understand relations, and extract structures, respectively. We conducted extensive evaluation of the effectiveness and improvement of our framework based on four description-based KGC models and four datasets, for both link prediction and triplet classification tasks.
\\ \newline  }

\begin{document}

\maketitleabstract

\section{Introduction}

A knowledge graph (KG) is a type of multi-relational graph data that contains the name/description of entities and relations and presents relational facts in a triplet format \cite{zhu2022multi}. Examples of KGs include Freebase~\cite{bollacker2008freebase}, DBpedia~\cite{lehmann2015dbpedia}, and YAGO~\cite{mahdisoltani2014yago3}, which have been proven useful in various applications, such as recommender systems \cite{sun2020multi} and knowledge graph question answer~\cite{saxena2020improving}.
Despite their widespread applications, KGs still suffer from the problem of incompleteness \cite{xu2023multimodal}. 
Along this line, the task of knowledge graph completion (KGC), which aims at predicting missing facts within a KG, helps both the construction and canonicalization of KGs. 

There have been several proposed works on KGC that aim to predict missing facts, such as structure-based KGC~\cite{sun2018rotate,dettmers2018conve,yue2023relation} and description-based KGC~\cite{wang-etal-2022-simkgc,wang2022language,jiang2023text}. The structure-based KGC only considers graph structural information from observed triple facts and embeds each entity and relation separately into trainable index embeddings. Unlike structure-based KGC, description-based KGC methods encode the text of entities and relations into a semantic space using pre-trained language models. The plausibility of facts is predicted by computing a scoring function of triplet or matching semantic similarity between the [head entity, relation] and tail entity~\cite{wang-etal-2022-simkgc}. In this way, the textual encoder facilitates easy generalization of the model to unseen graph entities, resulting in better scalability than index entity embedding.

\begin{figure}[]
\begin{center}
\includegraphics[scale=0.86]{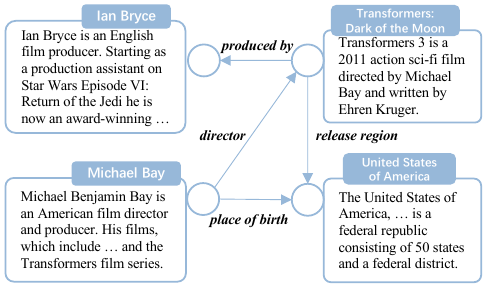} 
\caption{A subgraph with entity descriptions from a KG. The text of relations only includes its name.}
\label{fig.intro}
\end{center}
\end{figure}

However, despite the significant success achieved by description-based KGCs in learning textual and structural knowledge, their effectiveness is still limited by the quality of Internet-crawled text and incomplete structure. For instance, in Figure~\ref{fig.intro}, the brief descriptions of \textit{``Ian Bryce''} and \textit{``Transformers: Dark of the Moon''} are uninformative. In this case, relying solely on the name of relation \textit{``produced by''} may result in the ambiguous understanding of entity types. Meanwhile, learning structural patterns from known graphs is challenging for long-tailed entities. These limitations make it hard for KGC to achieve high performance in real-world applications that involve insufficient and incomplete knowledge graphs. 

Recently, large language models (LLMs) have been shown to have massive context-aware knowledge and advanced capability through the process of pre-training on large-scale corpora~\cite{liang2023holistic,peng2023large}. Therefore, it is worthwhile to utilize the rich knowledge of LLMs to address the challenges of KGC. However, it raises a question: \textit{how to effectively leverage the capabilities and knowledge of a language model to improve the graph learning?}

To answer the above question, in this paper, we propose a novel technical framework, called MPIKGC, which prompts LLMs to generate auxiliary texts for improving the performance of KGC models. Specifically, to address the problem of incomplete entity information, we propose to expand the entity description using the knowledge captured by LLMs. We achieve this by designing a Chain-of-Thought (CoT) prompt~\cite{wei2023chainofthought} that allows the LLM to break down the query into different aspects and generate descriptions step-by-step.
In addition to addressing the issue of relation ambiguity, we further propose a solution to improve KGC models' understanding of relation meanings, which involves querying LLMs with three curated prompting strategies, namely global, local, and reverse prompts that capture the association between relations and facilitate better reverse prediction.
Moreover, to address the issue of sparse graph links, especially for long-tailed entities, we propose enriching knowledge graphs by querying LLMs to extract additional structural information, using the keywords summarized by LLMs to measure similarity between entities, and creating new triples that construct associations between related entities and enable the formation of new structural patterns in KGC models.

To demonstrate the effectiveness and universality of our proposed framework, we apply the strategies from different perspectives to four description-based KGC models and four datasets separately, improving their performance on both the link prediction task and the triplet classification task. MPIKGC is also evaluated through a variety of ablation and comparison experiments, demonstrating its diversity for performance improvement from different perspectives and its generalizability across different LLMs. The codes and datasets are available in \url{https://github.com/quqxui/MPIKGC}.

%


\section{Related Work}

\subsection{Description-based KGC Methods}
Different from traditional structured-based KGC methods \cite{sun2018rotate}, which solely utilized structural information, description-based KGC methods typically represented entities and relations in KGs using pre-trained language models.
Specifically, these methods utilized textual descriptions for embedding entities and relations, and for the long-tailed entities, description-based KGC methods usually performed well because of the representation learning brought by entity descriptions.
For example, DKRL~\cite{xie2016representation} employed a convolutional neural network to encode entity descriptions, while KG-BERT~\cite{yao2019kgbert} utilized pre-trained BERT models to learn the embeddings of entities and relations, and KEPLER~\cite{wang2021kepler} further adapted pre-trained language models to simultaneously optimize knowledge embedding and language modeling objectives. MMRNS~\cite{MMRNS} leverages the similarity of description to perform negative sampling for hard samples.
Also, SimKGC, proposed by \citet{wang-etal-2022-simkgc}, utilized contrastive learning and various types of negative sampling to improve performance. LMKE~\cite{wang2022language}, on the other hand, leveraged language models and textual information to generate knowledge embeddings for entities, particularly for long-tailed entities. Besides, CSProm-KG~\cite{chen-etal-2023-dipping} extended frozen pre-trained language models with structural awareness using the proposed conditional soft prompt.

Despite the promising results of description-based KGC models, they still face difficulties caused by deficiency of textual data and incomplete structure, while our MPIKGC aims to address these challenges by introducing large language models to description-based KGC models.

\begin{figure*}[h]
\begin{center}
\includegraphics[scale=0.82]{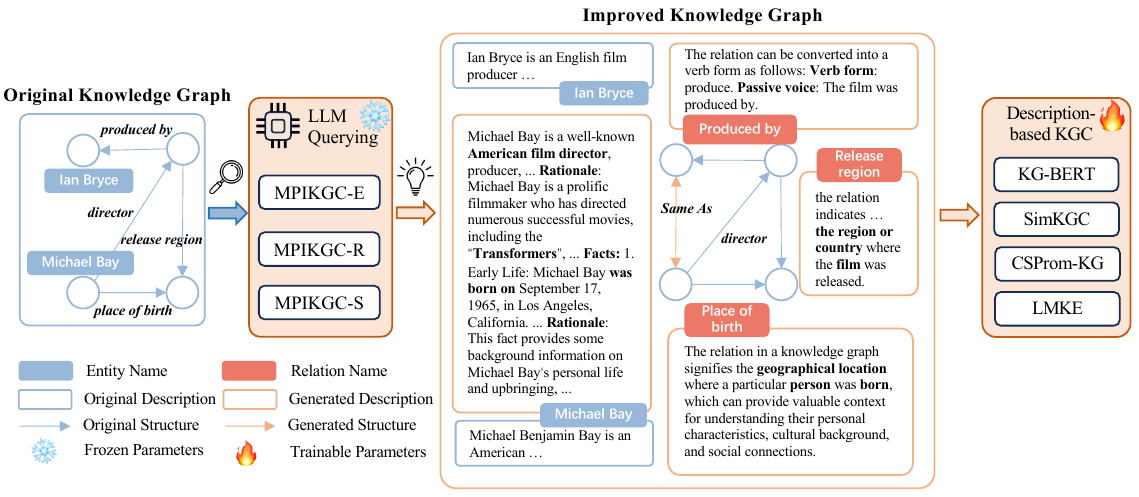} 
\caption{Framework of improving knowledge graph from the perspective of entity, relation, and structure. We evaluate the efficiency of the enhanced knowledge graph by employing various description-based KGC models on link prediction and triplet classification tasks. 
}
\label{fig.method}
\end{center}
\end{figure*}

\noindent
\subsection{Large Language Models for KG}
Recently, the emergence of large language models (LLMs), such as GPT4~\cite{openai2023gpt4}, Llama-2~\cite{touvron2023llama2}, and ChatGLM2~\cite{zeng2023glmb}, has led to several studies~\cite{pan2023large} exploring the potential of integrating LLMs and KGs to achieve improved performance by leveraging the strengths of both modalities. KGs can enhance LLMs by providing a means of explicitly storing rich factual knowledge, while LLMs can aid in the construction of KGs by generating new facts~\cite{pan2023unifying,xu2023large}.
Specifically, \citet{liang2023holistic} revealed that LLMs perform well on frequent entities and relations that mostly occur in the pre-training data.
The Chain-of-Thought (CoT)~\cite{wei2023chainofthought} prompting strategy significantly improved the reasoning performance of LLMs without requiring further fine-tuning.
Besides, \citet{mruthyunjaya2023rethinking} exhibited the considerable potential of LLMs in recalling factual information for symbolic KGs.
Moreover, LLMs have shown remarkable zero-shot performance in named entity recognition \cite{li2023prompt} as they can extract structural knowledge by utilizing relevant external context information.
Given these findings, we propose to investigate the potential of LLMs in terms of the reasoning, explanation, and summarization abilities for improving the KGC task.


\section{Methodology}


\begin{table*}[]
\renewcommand{\arraystretch}{1.1} 
\centering
\begin{tabular}{p{1.8cm}|p{12cm}}
\toprule
Strategies 	& Templates   \\
\midrule\midrule
MPIKGC-E & Please provide all information about \{Entity Name\}. Give the rationale before answering:   \\
\midrule
MPIKGC-R Global	& Please provide an explanation of the significance of the relation \{Relation Name\} in a knowledge graph with one sentence:   \\
\midrule
MPIKGC-R Local	& Please provide an explanation of the meaning of the triplet (head entity, \{Relation Name\}, tail entity) and rephrase it into a sentence:   \\
\midrule
MPIKGC-R Reverse	& Please convert the relation \{Relation Name\} into a verb form and provide a statement in the passive voice:   \\
\midrule
MPIKGC-S 	& Please extract the five most representative keywords from the following text: \{Entity Description\}. Keywords:   \\

\bottomrule
\end{tabular}
\caption{Templates of each strategy for querying. MPIKGC-E, -R, and -S are corresponding to the improvement methods in terms of entity, relation, and structure, respectively. }
\label{tab:template}
\end{table*}

\subsection{Problem Definition}
Knowledge graph $\mathcal{G}$ is a heterogeneous and direct graph data structure that can be represented as a collection of triplets with descriptions, which could be represented as  
$\mathcal{G} =  \{(h, r, t, d)\} \subseteq \mathcal{E} \times \mathcal{R} \times  \mathcal{E} \times  \mathcal{D} $, where $\mathcal{E}$ denotes the entity set, $\mathcal{R}$ as the relation set, and $\mathcal{D}$ as the original description of entities and relations. The aim of the triplet classification task is to ascertain the accuracy of a given triplet. The link prediction task of KGC is to infer the missing facts based on the known textual and structural data, which comprises two parts, i.e., predicting the tail entity when given $(h, r, ?) $, and predicting the head entity when provided with $(?, r, t)$. To accomplish this, it is necessary to rank all entities within $\mathcal{E}$ by calculating a score function for both positive and negative triples. Description-based KGC models utilize pre-trained language models to encode $\mathcal{D}$ and learn representations of entities and relations, while our goal is to enhance the textual and structural data as input of KGC models by querying LLMs with curated prompts.

\subsection{Multi-perspective Prompting}
In this section, we elaborate on the pipeline of our MPIKGC and provide motivation for each prompt, along with an illustrative example for better understanding. 
Our approach involves enhancing knowledge graph completion by improving entity, relation, and structure data, as shown in Figure~\ref{fig.method}. Specifically, templates for querying LLMs are shown in Table \ref{tab:template}, in which prompts used for querying LLMs follow three fundamental principles. (1) Clarity: It is crucial for LLMs to adhere to our instructions precisely. An excessively complex prompt may result in misunderstandings in the instructions, particularly for small LLMs (with fewer than 10 billion parameters), ultimately reducing the effectiveness of communication.
(2) Universality: The prompts we design should be compatible with various LLMs, and the generated text from these LLMs consistently demonstrates improvement on multiple KGC models, in both link prediction and triplet classification tasks.
(3) Diversity: Prompts should demonstrate diversity to enrich KG data from various perspectives: including entity, relation, and structure. They can improve the learning of KGC models and show cooperative effects when combined.
We evaluated the three claims of our proposed framework in the experiments presented in Section~\ref{sec:experiments}.

\subsubsection{Description Expansion}
Formalizing the comprehensive knowledge of an entity from LLMs is non-trivial, as it is difficult to ascertain whether the LLMs has generated and encompassed all the information of this entity.
Meanwhile, it can be challenging to manually set many instructions for each entity to query about, such as asking the released region or director for a movie, which consumes a lot of manpower and often results in an excessive number of tokens being input into the LLMs, consequently increasing the computational burden of inference. Such long text may also not be suitable for small-scale LLMs and can hamper their performance~\cite{bai2023longbench}.

We propose to design a Chain-of-Thought (CoT)~\cite{wei2023chainofthought} prompt strategy, that enables LLMs to break down complex queries into different directions and generate descriptions step-by-step, without the need for explicit manual input. It instructs LLMs to implicitly query relevant information on their own, resulting in more efficient and extensive responses.
As demonstrated template in Table~\ref{tab:template} \textbf{MPIKGC-E}, We request LLMs to provide a comprehensive entity description and provide a rationale before answering, which serves as justification for the answer and improves the recall of KGC models. For instance, Figure~\ref{fig.method} presents an example of a famous person \textit{``Michael Bay''}, where LLMs generate a description containing various occupations and personal details of the individual, accompanied by rationales for each response to enhance the LLMs statement.

\subsubsection{Relation Understanding}
The presence of heterogeneous relations in a knowledge graph plays a crucial role in distinguishing between two entities. However, relying solely on relation names may lead to ambiguous interpretations, particularly for complex relation categories (such as many-to-many and many-to-one). Moreover, the link prediction task requires an additional reverse prediction, i.e., predicting the head entity given  $(?, r, t)$. Typically, the performance of reverse prediction for many-to-one relation is significantly lower than that of forward prediction~\cite{yang2014embedding}.
Structure-based KGC methods attempt to address this issue by adding a reverse relation for each forward relation, thereby doubling the trainable index embeddings for relations.
In contrast, description-based KGC methods, such as SimKGC \cite{wang-etal-2022-simkgc}, append a string \textit{``reverse''} to the relation name. We argue that such an approach does not enable models to fully comprehend the meaning of relations, resulting in poor performance.

Therefore, we propose three prompting strategies, namely Global, Local, and Reverse, as depicted by MPIKGC-R in the Table~\ref{tab:template}. Specifically, \textbf{MPIKGC-R Global} aims to deduce the significance of a relation from the perspective of the entire KG, thereby facilitating better association between two relations. For instance, both \textit{``produced by''} and \textit{``director''} are related to the film industry, while \textit{``release region''} and \textit{``place of birth''} are associated with the name of a country or area.
In contrast, \textbf{MPIKGC-R Local} intends to infer the relation's meaning from the triplet perspective, thereby enhancing comprehension and suggesting possible types of head/tail entities while predicting missing facts. For instance, when querying the meaning of \textit{``(head entity, release region, tail entity)''}, LLMs suggest that the relation may connect to films and regions.
In addition, \textbf{MPIKGC-R Reverse} entails LLMs to represent relations as verbs, and convert them to the passive voice. For example, \textit{``produce''} can be transformed into \textit{``produced by''}, thereby enhancing comprehension and enabling better reverse prediction. The generated text is appended to the relation name and is processed according to each KGC model's workflow for handling the relation name.

\subsubsection{Structure Extraction}
KGC models are capable of learning structural patterns from training triples and generalizing to the missing links in test triples. For instance, a person entity that has the occupation of a producer or director, is probably related to film entities.
However, pattern learning from graph structures is limited to sparse links, particularly for long-tailed entities~\cite{li-etal-2022-learning-inter}. To address this problem, we propose \textbf{MPIKGC-S}, which queries LLMs to generate additional structural information to enrich KGs. To convert the LLMs' generative text into graph-based data, we utilize the summarizing capability of LLMs to extract relevant keywords from description, then calculate a matching score $s$ between entities based on the number of matched keywords:

\begin{equation*}
\begin{aligned}
    s &= len(m) /min(len(k_h),len(k_t)), \\
    m &= intersection(k_h,k_t), 
\end{aligned}
\end{equation*}
where $k_h$ and $k_t$ denote the keywords of head/tail entities, respectively, and $m$ is the intersection of $k_h$ and $k_t$.
After sorting the matching score, we selected top $k$ pairs and created new triples in the form of \textit{(head, Same As, tail)}, which are then appended to the training set. In addition to these similarity-based triplets, we also consider adding a self-loop triplet with the relation ``SameAs'' to each entity: \textit{(head, Same As, head)}. The motivation is to enhance the KGC model's learning of the ``SameAs'' relation.
These extra triplets construct the association between related entities and allow for the formation of new structural patterns in KGC models. For instance, by adding ``SameAs'' relation between \textit{``Ian Bryce''} and \textit{``Michael Bay''}, \textit{``Ian Bryce''} can reach to the \textit{``Transformers: Dark of the Moon''} entity with a explicit path, thereby serving as a valuable addition to the KGC model learning process.







\section{Experiments}\label{sec:experiments}

\subsection{Experimental Setup}

\begin{table}[]
\small
\setlength\tabcolsep{3pt}  
\renewcommand{\arraystretch}{1.1} 

\centering
\begin{tabular}{l|rrrrr}
\toprule
Dataset 	&\#Ent	&\#Rel		&\#Train &\#Valid &\#Test   \\
\midrule
FB15k237 	&14,541	&237		&272,115 &17,535 &20,466  \\
WN18RR 	&40,943	&11		&86,835 &3,034 &3,134  \\
FB13 &75,043	&13		&316,232 &5,908 &23,733  \\
WN11 &38,696&11&112,581&2,609&10,544\\
\bottomrule
\end{tabular}
\caption{Statistics of KGs used for our work.}
\label{tab:dataset}
\end{table}

\textbf{Datasets.} We conduct link prediction experiments on two widely used datasets, namely FB15k237~\cite{toutanova-chen-2015-observed} and WN18RR~\cite{dettmers2018conve}, as well as triplet classification experiments on FB13~\cite{socher2013reasoning} and WN11~\cite{socher2013reasoning}.

\noindent
\textbf{Metrics.} We evaluate the performance of KGC models using the following metrics: Mean Rank (MR), Mean Reciprocal Rank (MRR), and Hits@n (H@n, n=\{1,3,10\}) for the link prediction task, and Accuracy for the triplet classification task. Lower MR values indicate better performance, while higher values for other metrics are indicative of better performance.

\noindent
\textbf{Baselines.} In our study, we compare our improved KGs with the original KGs using four description-based KGC models: KG-BERT~\cite{yao2019kgbert}, SimKGC~\cite{wang-etal-2022-simkgc}, LMKE~\cite{wang2022language}, and CSProm-KG~\cite{chen-etal-2023-dipping}. The criteria for selecting baselines are based on state-of-the-art performance, the model's novelty, and the experimental time cost.
As KGC requires ranking all candidate entities, we prioritized baselines that utilized a 1-to-n scoring method.
%
We also compared against traditional structure-based KGC models, including TransE~\cite{bordes2013translating}, DistMult~\cite{yang2014embedding}, RotatE~\cite{sun2018rotate}, ConvE~\cite{dettmers2018conve}, ConvKB~\cite{nguyen2018novel}, and ATTH~\cite{chami-etal-2020-low}.

 \noindent
\textbf{Backbones.} We rely on Llama-2 (Llama-2-7b-chat)~\cite{touvron2023llama2}, ChatGLM2-6B~\cite{zeng2023glmb}, ChatGPT (gpt-3.5-turbo-0613)~\footnote{https://openai.com/blog/chatgpt}, GPT4 (gpt-4-0613)~\cite{openai2023gpt4} as our primary text generation backbones. In both LLMs, we set the temperature to 0.2 and the maximum length to 256, and use single-precision floating-point (FP32) for inference.
We employ BERT (bert-based-uncased)~\cite{yao2019kgbert} as the backbone to encode generated text for all description-based KGC models, with the hyperparameters being set in accordance with the corresponding models.

 \noindent
\textbf{Settings.} To ensure a fair comparison, we reproduce each method using their open-source codes and utilized the ``\texttt{bert-based-uncased}'' version~\cite{yao2019kgbert} as the backbone for all models. To account for the increased amount of text for both entities and relations in the enhanced KGs, we ensure that different augmentation experiments have the same maximal token length and data processing pipeline.
Additionally, we include Mean Rank results, which were not reported by some baselines.
Other parameters are set following the default parameters provided in the original paper.
To ensure the reproducibility of our results, we provide details on the hyper-parameters used for the four baselines across four benchmarks in Appendix \ref{sec:app_hyper}, which includes information on hyper-parameters such as maximum token length and batch size. Additional experiments are available in Appendix \ref{sec:app_Costs} and \ref{sec:app_TagReal}.

\subsection{Main Results}\label{sec:mainresults}

\begin{table*}[h]
\setlength\tabcolsep{2.8pt}  
\renewcommand{\arraystretch}{1.1} 

\centering
\begin{tabular}{l|rrrrr|rrrrr}
\toprule
\multicolumn{1}{c|}{\multirow{2}{*}{Models}}& \multicolumn{5}{c|}{FB15k237} & \multicolumn{5}{c}{WN18RR} \\
 \multicolumn{1}{c|}{}     & MR$\downarrow$& MRR$\uparrow$& H@1$\uparrow$ & H@3$\uparrow$ & H@10$\uparrow$& MR$\downarrow$& MRR$\uparrow$& H@1$\uparrow$ & H@3$\uparrow$ & H@10$\uparrow$\\ 
\midrule
 \multicolumn{11}{c}{\textit{Structure-based Approaches}} \\ 
\midrule
TransE{\scriptsize ~\cite{bordes2013translating}} & 323  &  27.9  & 19.8  & 37.6 & 44.1  & 2300 &24.3&4.3&44.1 & 53.2    \\
DistMult{\scriptsize ~\cite{yang2014embedding}} & 512  &  28.1  & 19.9  & 30.1 & 44.6  & 7000 &44.4 & 41.2  & 47.0 & 50.4    \\
ConvE{\scriptsize ~\cite{dettmers2018conve}} & 245  &  31.2  & 22.5  & 34.1 & 49.7  & 4464 &45.6&41.9&47.0 &53.1     \\
RotatE{\scriptsize ~\cite{sun2018rotate}} & 177  &  33.8  & 24.1  & 37.5 & 53.3  & 3340 &47.6&42.8&49.2 & 57.1    \\
ATTH{\scriptsize ~\cite{chami-etal-2020-low}} &  - &  34.8  & 25.2  & 38.4 & 54.0  & - &48.6&  44.3& 49.9& 57.3     \\
\midrule
 \multicolumn{11}{c}{\textit{Description-based Approaches}} \\
\midrule
CSProm-KG{\scriptsize ~\cite{chen-etal-2023-dipping}} &  188&	35.23	&26.05	&38.72	&53.57 & 545	&\textbf{55.10}	&\textbf{50.14}	&\textbf{57.04}	&64.41  \\
\quad+MPIKGC-E&  195	 &35.51	 &26.38 &	38.96 &	53.74   & 1244 &53.80 &	49.19 &	55.65	 &62.81  \\
\quad+MPIKGC-R&  192&	35.38	&26.29	&38.83	&53.50  &838&	53.90&	49.35	&55.74	&62.36     \\
\quad+MPIKGC-S&  \textbf{179}	&\textbf{35.95}	&\textbf{26.71}&	\textbf{39.52}	&\textbf{54.30}  &\textbf{528}	& 54.89	&49.65	&56.75	&\textbf{65.24}    \\
\midrule\midrule
LMKE{\scriptsize ~\cite{wang2022language}} & \textbf{135} & 	30.31& 	21.49& 	33.02& 	48.07  & \textbf{54} &55.78& 	42.91	& 64.61	& 79.28    \\
\quad+MPIKGC-E&       138	&30.83&	21.89&	33.67	&48.75     &  57	& 56.35& 	43.27& 	65.54	& \textbf{79.53}    \\
\quad+MPIKGC-R&   145	&\textbf{30.99}&	\textbf{22.21}	&\textbf{33.70}	&48.83&  59	& \textbf{57.60}	& \textbf{45.10}	& \textbf{65.95}	& 79.35     \\
\quad+MPIKGC-S&   \textbf{135}& 	30.68& 	21.67& 	33.35	& \textbf{48.91}   &70&	50.71&	36.91	&59.65	&76.13    \\
\midrule\midrule
SimKGC{\scriptsize ~\cite{wang-etal-2022-simkgc}} &  146 & 32.66   & 24.13  & 35.42 &  49.65 & 148 &65.64 &57.08 &71.20 & 80.33    \\
\quad+MPIKGC-E&  \textbf{143}&	33.01&	24.37&	35.80&	50.29 &   \textbf{ 124}&	65.64&	57.10&	71.09&	80.41   \\
\quad+MPIKGC-R& 156&31.05&	22.63&	33.62&	47.65 &  129	 &\textbf{66.41}	 &\textbf{57.90}	 &\textbf{72.08}	 &\textbf{81.47}    \\
\quad+MPIKGC-S&\textbf{143}&	\textbf{33.22}&	\textbf{24.49}	&\textbf{36.26}&	\textbf{50.94} &  170	&61.48	&52.81	&66.77	&76.94    \\
\bottomrule
\end{tabular}
\caption{Experimental results on the link prediction task where the best results in each block are \textbf{in bold}. $\uparrow$: higher is better. $\downarrow$: lower is better.}
\label{tab:linkprediction}
\end{table*}

In this section, we conduct a comprehensive performance comparison between structure-based KGC and description-based KGC and evaluate the effectiveness of our MPIKGC in three perspectives by feeding the enhanced KGs to description-based approaches.
As presented in Table~\ref{tab:linkprediction}, ATTH outperforms other structure-based methods across four metrics. Besides, we observe that in most situations, structure-based methods exhibited better performance on FB15k237, which contains general world facts (such as an entity of actor), while description-based methods perform better on WN18RR, a subset of WordNet~\cite{miller1995wordnet} with rich language knowledge suitable for PLMs.

CSProm-KG proposed to focus on both textual and structural information, which results in its superior performance on FB15k237 compared to LMKE and SimKGC. However, it performs much worse on WN18RR.
Our proposed methods, particularly the structure extraction approach MPIKGC-S, improve the structure-focused aspect of CSProm-KG and achieve the highest performance on FB15k237 compared to all other baselines, even surpassing the structure-based methods. However, the difference between the FB15k237 and WN18RR datasets is indeed noteworthy. FB15k237 has shown particularly good results with MPIKGC-S, which could be attributed to the fact that FB15k237 has 15K entities and 237 kinds of relations as shown in Table \ref{tab:dataset}, while WN18RR has 40K entities with only 11 relations. Adding extra relation to WN18RR might excessively alter the sparsity and triplet distribution of the KG, leading to poor performance.

On the other hand, our proposed method for relation understanding (MPIKGC-R) demonstrates a 1\%-2\% improvement in the MRR, Hits@1, and Hits@3 metrics on WN18RR compared to LMKE, while MPIKGC-E achieves higher Hits@10 score of 79.53\%. The same improvement trend is also observed for the MPIKGC-R approach when applied to SimKGC on WN18RR. The reason is that both methods are text-focused and have the capability to learn abundant information with enhanced data. However, MPIKGC-S does not exhibit improvement in LMKE and SimKGC on WN18RR. We hypothesize that this may be attributed to the low number of relation types, which could potentially mislead the model when adding new relations.
Moreover, the incorporation of extracted structural data for FB15k237 achieves much better performance based on LMKE and SimKGC, which proves the effectiveness of MPIKGC-S in the link prediction.

\subsection{Triplet Classification}

\begin{table}[t]
\renewcommand{\arraystretch}{1.1} 

\centering
\begin{tabular}{l|ccccc|ccccc}
\toprule
Models & FB13  &  WN11 \\
\midrule
 \multicolumn{3}{c}{\textit{Structure-based Approaches}} \\ 
\midrule
   TransE{\scriptsize ~\cite{bordes2013translating}} &  81.5 &  75.9  \\
DistMult{\scriptsize ~\cite{yang2014embedding}} & 86.2  & 87.1   \\
ConvKB{\scriptsize ~\cite{nguyen2018novel}} &88.8& 87.6\\
\midrule
 \multicolumn{3}{c}{\textit{Description-based Approaches}} \\ 
\midrule
KG-BERT{\scriptsize ~\cite{yao2019kgbert}} & 84.74  &  93.34  \\
\quad+MPIKGC-E& \textbf{86.29}  &    \textbf{94.13}    \\
\quad+MPIKGC-R& 84.51  &   93.36  \\
\quad+MPIKGC-S&  85.35 &    93.61  \\
\midrule\midrule
LMKE{\scriptsize ~\cite{wang2022language}} & 91.70  & 93.71  \\
\quad+MPIKGC-E&  91.52 &     93.84   \\
\quad+MPIKGC-R& 91.49  &   \textbf{93.93}  \\
\quad+MPIKGC-S& \textbf{91.81}  &  93.91    \\
\bottomrule
\end{tabular}
\caption{Accuracy on the triplet classification task.}
\label{tab:classification}
\end{table}

In this section, we evaluate our proposed methods on the triplet classification task, a binary classification task that determines the correctness of a given triplet.
Based on the results presented in Table~\ref{tab:classification}, we can conclude that the structure-based methods perform well on the FB13 dataset, while significantly underperform compared to the description-based methods on the WN11 dataset. This outcome is consistent with the findings in the link prediction task discussed in Section~\ref{sec:mainresults} and can be attributed to the variations between Freebase~\cite{bollacker2008freebase} and WordNet~\cite{miller1995wordnet}. On the other hand, the results demonstrate that expanding descriptions (MPIKGC-E) is a promising technique to improve the KG-BERT's performance, as it yields a 1.55\% higher accuracy score on FB13 and a 0.79\% higher accuracy score on WN11. Our methods also exhibit minor enhancements on LMKE and achieve the highest accuracy score of 91.81\% in FB13.
The overall results indicate the universality of the MPIKGC framework that can enhance the performance of various KGC models in both link prediction and triplet classification tasks.

\begin{figure*}[h]
\centering
\includegraphics[width=0.5\columnwidth]{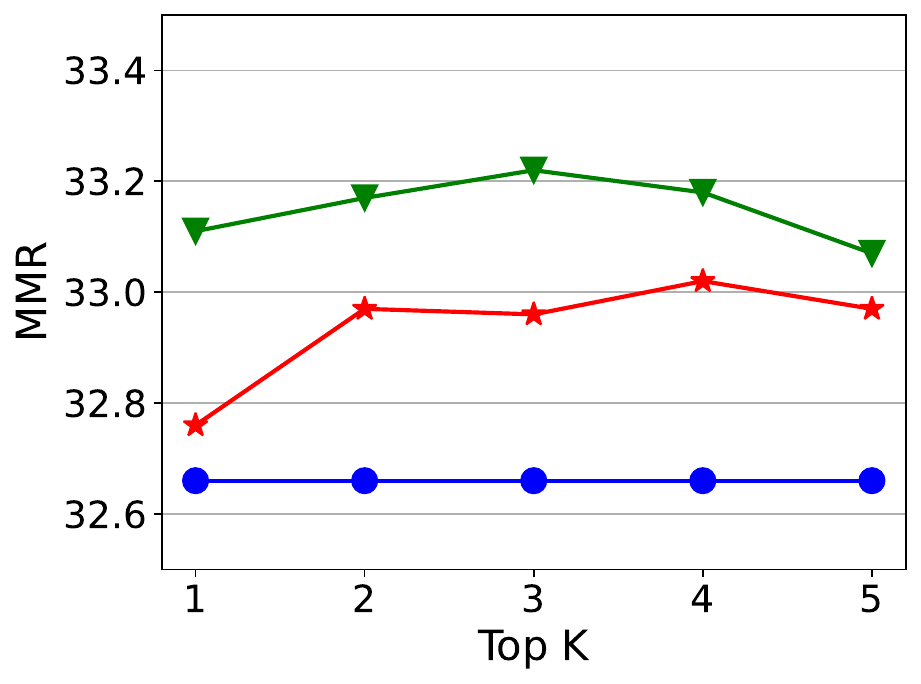}
\includegraphics[width=0.5\columnwidth]{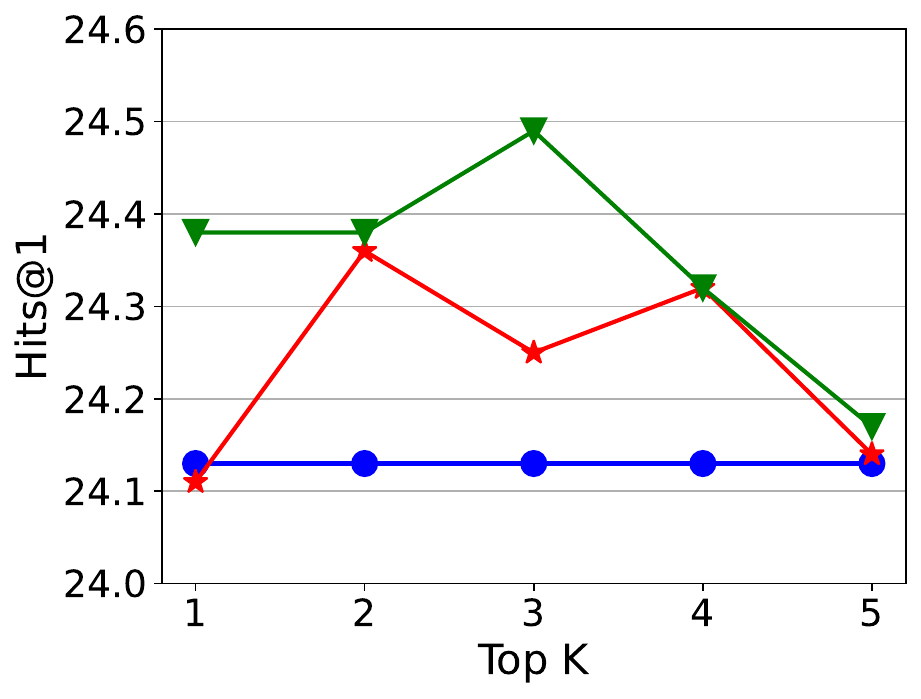}
\includegraphics[width=0.5\columnwidth]{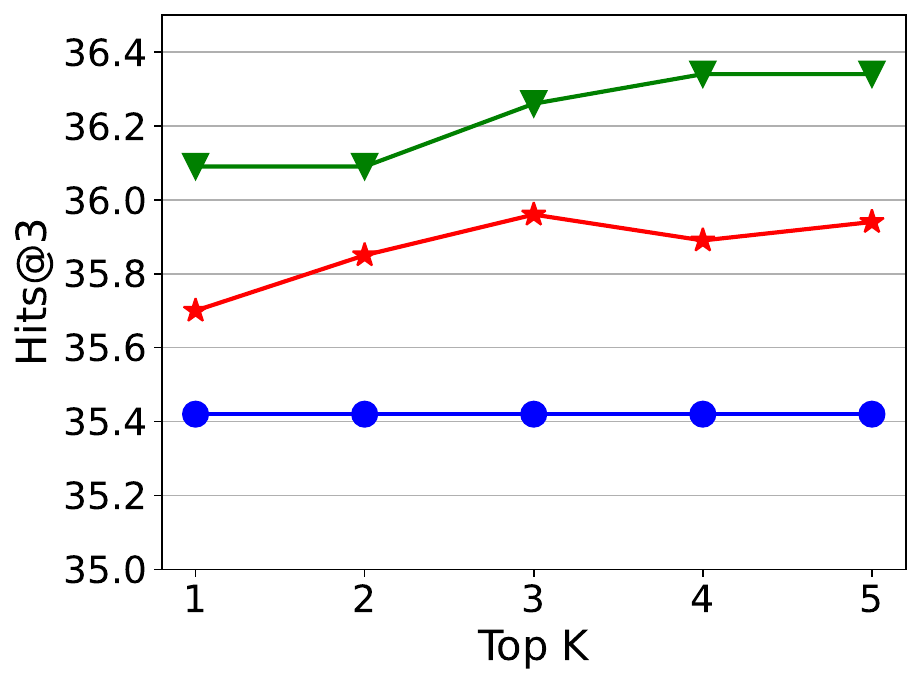}
\includegraphics[width=0.5\columnwidth]{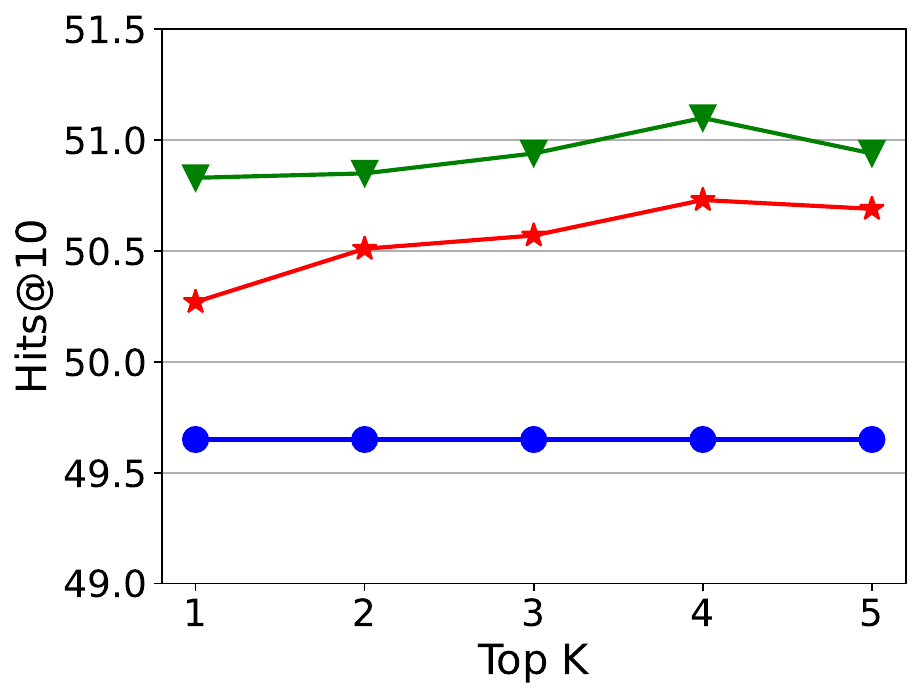}
\includegraphics[width=0.9\columnwidth]{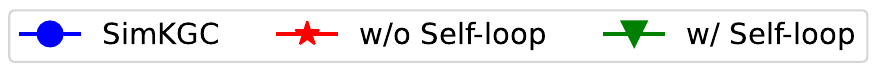}

\caption{Analysis of hype-parameter $k$ and the self-loop setting on FB15k237. }
\label{fig:struc}
\end{figure*}

\subsection{Parameter Analysis of Structure Extraction}

In this section, we evaluate the significance of the hyper-parameter top $k$ in extracting structural data. We present the results obtained on FB15k237 and compare the settings with and without self-loop with SimKGC. Increasing the value of $k$ implies the addition of more triplets to the training set. For instance, when $k$ is set to 1, we identify the best-matched entity for each entity by computing the matching score. Besides, the self-loop setting involves the inclusion of a triplet for the entity itself.
As shown in Figure~\ref{fig:struc}, the curves for all three metrics exhibit an upward trend under both settings as the $k$ increases, which demonstrates that the augmentation perspective in structure indeed improves the performance of KGC on the FB15k237 dataset. Furthermore, we can see that the setting with self-loop consistently outperforms the setting without self-loop. This observation suggests that augmenting KGC's learning of the ``SameAs'' relation is a promising strategy for enhancing performance.
However, we also note that the performance starts to decline when $k$ reaches 4 or 5. This trend indicates that the training set contains excessive extraneous triplets, which could negatively affect the learning of other data.

\begin{table}[t]
\small
\setlength\tabcolsep{5pt}  
\renewcommand{\arraystretch}{1.1} 

\centering
\begin{tabular}{l|cccc}
\toprule
\multicolumn{1}{l|}{\multirow{2}{*}{Models}} & \multicolumn{4}{c}{FB15k237} \\
 \multicolumn{1}{c|}{}    & MRR& H@1 & H@3  & H@10\\ 
\midrule\midrule
LMKE&30.31 &21.49& 33.02 &48.07	    \\
+MPIKGC-E		&30.71	&21.97	&33.29	&48.35 \\ 
+MPIKGC-R 	&30.64&	21.70	&33.22&	48.74    \\
+MPIKGC-S &30.68	&21.67	&33.35	&48.91\\
+MPIKGC-E\&R	&30.74	&21.77	&33.57	&48.77   \\
+MPIKGC-E\&S 	 	&30.92	&21.85 &33.67	&\textbf{49.50}  \\
+MPIKGC-R\&S 		&\textbf{31.21}	&\textbf{22.26}	&33.86	&49.42  \\
+MPIKGC-E\&R\&S	&30.97	&21.91 &\textbf{33.90}	&49.28 \\
\bottomrule
\end{tabular}
\caption{Ablation of augmentation methods from different perspectives.}
\label{tab:Ablation-ERS}
\end{table}

\subsection{Ablation of Multi-perspective Prompts}

This section presents an analysis of the performance of description expansion (MPIKGC-E), relation understanding (MPIKGC-R), and structure extraction (MPIKGC-S), as well as the performance when combining them together.
For instance, MPIKGC-E\&R denotes the combination of MPIKGC-E and MPIKGC-R, while the remaining methods follow the same naming convention. We conduct ablation experiments on the FB15k237 dataset using Llama-2-generated texts, with LMKE as the baseline. The results demonstrate that after being enhanced with our method from the perspectives of entities, relations, and structures, the KGC models achieved a nearly 0.5\% improvement across all four metrics. Additionally, MPIKGC-E\&R combines the generated entity descriptions with the descriptive text of relations, resulting in a slight improvement over using either one individually, which demonstrates the compatibility of these two methods. Moreover, MPIKGC-E\&S achieves the highest H@10 score, while MPIKGC-R\&S performs best on MRR and H@1. MPIKGC-E\&R\&S achieves the best H@3 score. We observe that incorporating the structure extraction method further improves performance by nearly 0.5\% across most metrics. For example, MPIKGC-E\&R\&S gets a H@10 score of 49.28\%, which is 0.51\% higher than MPIKGC-E\&R. The same phenomenon can be seen when adding relation text `-R'. The comprehensive findings indicate that our diverse enhancement methods are compatible and can be integrated to boost overall performance.

\subsection{Ablation of Relation Understanding}

In this section, we evaluate the performance of various ablation settings for relation understanding on the WN18RR dataset, as shown in Table~\ref{tab:ablation_relation}. 

Specifically, MPIKGC-R G\&L represents the combination of Global and Local descriptions, which are concatenated using a separate token `[SEP]'. Meanwhile, the other methods follow the same rule. The results show that MPIKGC-R Global outperformed the baseline SimKGC by nearly 1\% across all four metrics.
Additionally, MPIKGC-R Local achieves the highest H@10 score of 81.57\%, but has the lowest MRR and H@1 scores. Conversely, MPIKGC-R Reverse achieves over 1\% improvement in MRR and H@1 but performs worse in H@10.
These results suggest that MPIKGC-R Local prioritizes the top-10 recall of correct entities, while MPIKGC-R Reverse focuses on improving the performance of the best entity (i.e., top-1).

After combining these three strategies, we observe that MPIKGC-R G\&L achieves a significant improvement in MRR and H@1, suggesting that Global and Local prompts have a complementary effect. However, other combined strategies exhibit poor performance. We therefore believe that incorporating too many relation descriptions may increase the difficulty of learning the meaning of the relation.

\begin{table}[t]
\small
\setlength\tabcolsep{4pt}  
\renewcommand{\arraystretch}{1.1} 
\centering
\begin{tabular}{l|cccc}
\toprule
\multicolumn{1}{l|}{\multirow{2}{*}{Models}} & \multicolumn{4}{c}{WN18RR} \\
 \multicolumn{1}{c|}{}    & MRR& H@1 & H@3 & H@10\\ 
\midrule\midrule
SimKGC 	&65.64	&57.08		&71.20&80.33    \\
+MPIKGC-R Global	&66.41	&57.90	&	\textbf{72.08}&81.47  \\
+MPIKGC-R Local&	64.45	&54.87 &70.65 &	\textbf{81.57}        \\
+MPIKGC-R Reverse	&66.53&	59.28	&	70.72&80.09    \\
+MPIKGC-R G\&L 	&\textbf{66.97}	&\textbf{59.88}	&70.82&79.77     \\
+MPIKGC-R G\&R	&65.56	&57.00&70.98&	80.90   \\
+MPIKGC-R L\&R	&65.75&	57.36	&71.03&	80.06    \\
+MPIKGC-R G\&L\&R&	65.85 &	57.47	&70.98&	80.64\\
\bottomrule
\end{tabular}
\caption{Ablation of different relation understanding strategies and combinations on WN18RR.}
\label{tab:ablation_relation}
\end{table}

\subsection{Comparison of LLMs}\label{sec:Comparison of LLMs}

This section is dedicated to exploring the use of various LLMs to enhance KGC on FB15k237. Due to the long querying time and high costs associated with querying each entity and keyword on four benchmarks, we have restricted our analysis to the application of ChatGPT and GPT4 solely for MPIKGC-R. 
As shown in Table~\ref{Ablation-llm}, the results indicate that our framework consistently improves KGC based on LMKE with generated text across three perspectives when using various LLMs. This suggests the effectiveness of our designed prompts, which are universal to both large-scale (ChatGPT and GPT4) and small-scale (Llama-2 and ChatGLM2) LLMs.
Specifically, ChatGLM2 produces superior results for MPIKGC-E and MPIKGC-S compared to Llama-2, indicating ChatGLM2's advantage of reasoning and summarization abilities. However, Llama-2 and ChatGPT outperform ChatGLM2 in terms of their ability to understand relations.
On the other hand, we can see that applying GPT4 into MPIKGC-R has led to a significant improvement in all metrics, owing to the larger model scale that facilitates a more comprehensive understanding of KG relations.

\begin{table}[t]
\small
\setlength\tabcolsep{4pt}  
\renewcommand{\arraystretch}{1.1} 
\centering
\begin{tabular}{l|cccc}
\toprule
\multicolumn{1}{l|}{\multirow{2}{*}{Models}} & \multicolumn{4}{c}{FB15k237} \\
\multicolumn{1}{c|}{}   & MRR& H@1 & H@3 & H@10 \\
\midrule
LMKE & 	30.31& 	21.49& 	33.02& 	48.07  \\
\midrule
{\scriptsize +MPIKGC-E (Llama-2)} 	&30.56&	21.62	&33.47	&48.15  \\
{\scriptsize +MPIKGC-E (ChatGLM2)} 	&\textbf{30.83}&	\textbf{21.89}&	\textbf{33.67}	&\textbf{48.75} \\
\midrule
{\scriptsize +MPIKGC-R (Llama-2)}   	&30.64&	21.70	&33.22	&48.74 \\
{\scriptsize +MPIKGC-R (ChatGLM2)} 	&	30.24	&21.33	&32.96	&48.27\\
{\scriptsize +MPIKGC-R (ChatGPT)}    & 	30.65 &	21.82	 &33.24 &	48.52   \\
{\scriptsize +MPIKGC-R (GPT4)}   	&\textbf{30.99}	&\textbf{22.21}	&\textbf{33.70}	&\textbf{48.83}   \\
\midrule
{\scriptsize +MPIKGC-S (Llama-2)}   &30.68	&21.67	&33.35	&\textbf{48.91}\\
{\scriptsize +MPIKGC-S (ChatGLM2)}      &	\textbf{31.07}	&	\textbf{22.26}	&	\textbf{33.81}	&	48.82 \\
\bottomrule
\end{tabular}
\caption{Ablation of different LLMs on FB15k237.}
\label{Ablation-llm}
\end{table}

\section{Conclusion}
In this paper, we proposed MPIKGC, a novel framework that investigates improving the quality of KGs by querying LLMs from three perspectives: expanding the entity descriptions by designing Chain-of-Thought prompt, enhancing the understanding of relation by designing global, local, and reverse prompts, as well as extracting the structural data via keywords summarization and matching. 

We evaluated MPIKGC on WN18RR and FB15k237 datasets for the link prediction task and on WN11 and FB13 datasets for the triplet classification task. The results of extensive experiments demonstrated that our method achieved significant improvement over four description-based KGC models. Moreover, additional ablation experiments highlight the potential to combine different enhancement methods for even better performance.

In the future, we plan to explore the possibility of refining the "SameAs" relation into more fine-grained categories, without adding too many triplets. On the other hand, generating KG data by LLMs may encounter problems such as hallucination, toxic and bias, and we plan to develop restricted prompts or fine-tune LLMs to augment text generation controllability and interpretability.

\section{Acknowledgement}
This work was supported in part by the grants from National Natural Science Foundation of China (No.62222213, U22B2059, U23A20319, 62072423), and the USTC Research Funds of the Double First-Class Initiative (No.YD2150002009). Xiangyu Zhao was partially supported by Research Impact Fund (No.R1015-23), APRC - CityU New Research Initiatives (No.9610565, Start-up Grant for New Faculty of CityU), CityU - HKIDS Early Career Research Grant (No.9360163), Hong Kong ITC Innovation and Technology Fund Midstream Research Programme for Universities Project (No.ITS/034/22MS), Hong Kong Environmental and Conservation Fund (No. 88/2022), and SIRG - CityU Strategic Interdisciplinary Research Grant (No.7020046, No.7020074), and CCF-Tencent Open Fund.

\balance
\section{Bibliographical References}\label{sec:reference}

\bibliographystyle{template_files/lrec-coling2024-natbib}
\bibliography{main_bib}


\section{Appendices}\label{sec:apps}
\subsection{Hyper-Parameter Settings}\label{sec:app_hyper}
To ensure the reproducibility of our results, we provide detailed information on the hyperparameters used for the four baselines across four benchmarks. This includes information on hyperparameters such as max token length and batch size, as shown in Table \ref{tab:HP-CSProm}, \ref{tab:HP-simkgc}, \ref{tab:HP-KG-BERT}, and \ref{tab:HP-lmke}.

To ensure a fair evaluation of the performance improvement of our enhanced KG on KG completion compared to the original KG, we maintain the same hyperparameters for them. The parameter settings for the four KGC models mostly follow the defaults in their open-source code.

\begin{table}[H]
\centering
\begin{tabular}{l|cc}
\toprule
KG-BERT & FB13 &  WN11 \\
\midrule
learning rate  &  5e-5 & 5e-5 \\
batch size  & 32 & 256 \\
epochs & 8  & 5 \\
max num tokens & 70  & 50 \\
gradient accumulation steps &1& 1\\
warmup proportion &0.1  & 0.1 \\
\bottomrule
\end{tabular}
\caption{The hyper-parameter settings of KG-BERT for FB13 and WN11.}
\label{tab:HP-KG-BERT}
\end{table}

 \begin{table}[H]
\centering
\begin{tabular}{l|cc}
\toprule
SimKGC & FB15k237 &  WN18RR \\
\midrule
learning rate  &  1e-5 & 5e-5 \\
batch size  & 1024 & 1024 \\
additive margin & 0.02 & 0.02 \\
use amp & True  & True \\
use self-negative & True  & True \\
finetune-t & True  & True \\
pre-batch & 2  & 0 \\
epochs & 10  & 50 \\
max num tokens & 70  & 50 \\
\bottomrule
\end{tabular}
\caption{The hyper-parameter settings of SimKGC for FB15k237 and WN18RR.}
\label{tab:HP-simkgc}
\end{table}

 \begin{table}[H]
\centering
\begin{tabular}{l|cc}
\toprule
CSProm-KG & FB15k237 &  WN18RR \\
\midrule
learning rate  &  5e-4 & 5e-4  \\
batch size  & 128 & 128 \\
epochs & 60  & 60 \\
desc max length & 70  & 50 \\
prompt length &10 &10\\
alpha &0.1& 0.1\\
n\_lar &8 &8\\
label smoothing& 0.1&0.1\\
embed dim &156& 144\\
k\_w &12 &12\\
k\_h &13& 12\\
alpha step& 0.00001 &0.00001\\
\bottomrule
\end{tabular}
\caption{The hyper-parameter settings of CSProm-KG for FB15k237 and WN18RR.}
\label{tab:HP-CSProm}
\end{table}

 \begin{table}[H]
\small
\setlength\tabcolsep{3pt}  
\centering
\begin{tabular}{l|cc}
\toprule
LMKE & FB15k237/FB13 &  WN18RR/WN11 \\
\midrule
bert\_lr  &  1e-5 & 1e-5 \\
model\_lr &5e-4 &5e-4\\
weight decay &1e-7 & 1e-7\\
batch size  & 256 & 1024 \\
epochs & 8  & 70 \\
max tokens & 70  & 50 \\
self adversarial &True &True\\
contrastive & True&True\\
plm & bert & bert\\
\bottomrule
\end{tabular}
\caption{The hyper-parameter settings of LMKE for FB15k237, FB13, WN11, and WN18RR.}
\label{tab:HP-lmke}
\end{table}

\subsection{Costs}\label{sec:app_Costs}
For the LLM inference stage, we conduct experiments on one NVIDIA V100 32G, and report the average time and config/cost for 100 data samples, as shown in Table \ref{tab:MPIKGC-E}, \ref{tab:MPIKGC-R}, and \ref{tab:MPIKGC-S}.
For example, it only costs 11.25 hours of inference time to run MPIKGC-E on FB15k237 with 15k samples, using one V100 32G.
For the KGC training stage, our enhancement method requires minimal additional time over the original method, with only an increase of 2G-3G in GPU memory consumption due to the larger maximum token.

 It is important to note that the majority of our experiments do not rely on ChatGPT API. Instead, we primarily utilize open-source LLMs such as Llama and ChatGLM (in the main experiments). ChatGPT API is only involved in Section \ref{sec:Comparison of LLMs} for the ablation study of comparing different LLMs.
 \begin{table}[H]
\centering
\begin{tabular}{l|cc}
\toprule
LLM & config/cost &  time \\
\midrule
chatglm-2-6b  & V100 13G & 5.4s \\
llama-2-7b-chat  & V100 14G & 6.9s \\
gpt-3.5-turbo-0613 & 0.000638\$ & 4.9s \\
gpt-4-0613 & 0.0246\$  & 13.0s \\
\bottomrule
\end{tabular}
\caption{Generation costs for MPIKGC-E.}
\label{tab:MPIKGC-E}
\end{table}

 \begin{table}[H]
\centering
\begin{tabular}{l|cc}
\toprule
LLM & config/cost &  time \\
\midrule
chatglm-2-6b  & V100 13G & 3.1s \\
 llama-2-7b-chat  & V100 14G & 3.3s \\
 gpt-3.5-turbo-0613 & 0.00021\$  & 2.8s \\
 gpt-4-0613  & 0.00615\$  & 3.9s \\
\bottomrule
\end{tabular}
\caption{Generation costs for MPIKGC-R.}
\label{tab:MPIKGC-R}
\end{table}

 \begin{table}[H]
\centering
\begin{tabular}{l|cc}
\toprule
LLM & config/cost &  time \\
\midrule
 chatglm-2-6b  & V100 13G & 3.4s \\
 llama-2-7b-chat  & V100 14G & 3.3s \\
 gpt-3.5-turbo-0613 & 0.00020\$ & 2.7s \\
 gpt-4-0613  & 0.00318\$  & 5.2s \\
\bottomrule
\end{tabular}
\caption{Generation costs for MPIKGC-S.}
\label{tab:MPIKGC-S}
\end{table}

\subsection{Comparison with TagReal}\label{sec:app_TagReal}
Here, we compare TagReal \cite{jiang2023text} to demonstrate our differences and advantages for knowledge graph completion.
TagReal is a method for open knowledge graph completion that automatically generates query prompts and retrieves support information from text corpora to probe knowledge from pre-trained language models. 
Theoretically, while TagReal performs well on KGC, we identify several improvements in our method compared to Targal:
\begin{itemize}
    \item We propose three enhancement strategies that improve performance from different perspectives (entity, relation, structure). Each strategy has its own advantages and, when combined, they further enhance performance.
    \item Our method exhibits stronger applicability. Our method can enhance most description-based KGC models, adapt to multiple LLMs, and improve the performance of both link prediction and triplet classification tasks. 
    \item Our method is more flexible as it generates supplementary text descriptions using LLMs through instructions. However, the prompt generation of TagReal requires complex steps. We believe that our designing instructions and generating prompts using LLMs would be simpler and more efficient.
\end{itemize}
 As for experimental comparison in Table \ref{tab:TagReal}, the method TagReal differs from ours in terms of the benchmarks, backbones, and training methods used. We compared the prompt generated for each relation from TagReal with our method MPIKGC-R in FB15k237. For instance, for the relation "\textit{/sports/sports\_team\_location/teams}", the enhancement strategies are as follows:
\begin{itemize}
    \item TagReal --> [X] is located in [Y].
    \item MPIKGC-R --> The relation denotes the geographical location in which sports teams are located, indicating the connection between teams and their respective locations.
\end{itemize}

 \begin{table}[H]
\small
\setlength\tabcolsep{5pt}  
\renewcommand{\arraystretch}{1.1} 

\centering
\begin{tabular}{l|cccc}
\toprule
\multicolumn{1}{l|}{\multirow{2}{*}{Models}} & \multicolumn{4}{c}{FB15k237} \\
 \multicolumn{1}{c|}{}    & MRR& H@1 & H@3  & H@10\\ 
\midrule\midrule
LMKE&30.31 &21.49& 33.02 &48.07	    \\
+TagReal		&30.48 & 21.63 &33.19  & 48.37 \\ 
+MPIKGC-R Global	&30.99  &22.21  & 33.70  &48.83  \\
\bottomrule
\end{tabular}
\caption{Comparison with TagReal, in which the enhancement data for MPIKGC-R Global was generated by querying GPT4.}
\label{tab:TagReal}
\end{table}

The experimental results show that after supplementing the relation, our method can generate a more detailed explanation for this relation, resulting in better performance, even though TagReal shows performance improvement compared to LMKE.

\end{document}